\documentclass{llncs}

\usepackage[utf8]{inputenc}
\usepackage{amsmath}
\usepackage{graphicx}
\usepackage{amsfonts}
\usepackage{amssymb}
\usepackage{url}
\usepackage{textgreek}
\usepackage{booktabs}

\usepackage[english]{babel}
\usepackage[nottoc]{tocbibind}

\usepackage{todonotes}
\usepackage{color}

\usepackage[normalem]{ulem}

\begin{document}

{\let\thefootnote\relax\footnotetext{Copyright \textcopyright\ 2020 for this paper by its authors. Use permitted under Creative Commons License Attribution 4.0 International (CC BY 4.0). CLEF 2020, 22-25 September 2020, Thessaloniki, Greece.}}

\title{Team\_Alex at CLEF CheckThat! 2020: Identifying Check-Worthy Tweets\\ With Transformer Models}

\author{%
Alex Nikolov\inst{1}
\and
Giovanni Da San Martino\inst{2}
\and
Ivan Koychev\inst{1}
\and
Preslav Nakov\inst{2}
}

\institute{%
FMI, Sofia University ``St Kliment Ohridski'', Bulgaria
\and
Qatar Computing Research Institute, HBKU, Qatar
}

\maketitle

\begin{abstract}
While misinformation and disinformation have been thriving in social media for years, with the emergence of the COVID-19 pandemic, the political and the health misinformation merged, thus elevating the problem to a whole new level and giving rise to the first global infodemic. The fight against this infodemic has many aspects, with fact-checking and debunking false and misleading claims being among the most important ones. Unfortunately, manual fact-checking is time-consuming and automatic fact-checking is resource-intense, which means that we need to pre-filter the input social media posts and to throw out those that do not appear to be check-worthy.
With this in mind, here we propose a model for detecting check-worthy tweets about COVID-19, which combines deep contextualized text representations with modeling the social context of the tweet. 
Our official submission to the English version of CLEF-2020 CheckThat! \mbox{Task 1}, system Team\_Alex, was ranked second with a MAP score of 0.8034, which is almost tied with the wining system, lagging behind by just 0.003 MAP points absolute.
\end{abstract}

\section{Introduction}
\label{sec:introduction}

The rise of disinformation (aka ``fake news'') in social media has given rise to a number of initiatives to fact-check claims of general interest and to confirm or to debunk them. Unfortunately, manual fact-checking is a very time-consuming process, and thus automated approaches have been proposed as a faster alternative. Yet, even with automated methods, it is not possible to fact-check every single claim, and the accuracy of automated fact-checking systems is significantly lower than that of human experts. Furthermore, there is a need to pre-filter and to prioritize what should be passed to human fact-checkers. As the need to prioritize has been gradually gaining recognition, so was the task of check-worthiness estimation, which is seen as an important first step in the general fact-checking pipeline. A leading effort in this direction has been the CLEF CheckThat! lab, which featured a check-worthiness estimation task in all its editions \cite{clef2018checkthat:task1,clef-checkthat-T1:2019,clef-checkthat-en:2020}.

Traditionally, check-worthiness estimation has focused on political debates and speeches, ignoring social media. In order to bridge this gap, the 2020 edition of the CLEF-2020 CheckThat! Lab~\cite{clef-checkthat:2020} 
featured Task 1 on check-worthiness estimation on tweets, offered in Arabic and English~\cite{clef-checkthat-ar:2020,clef-checkthat-en:2020}. 
Given the prominence of disinformation related to COVID-19, which has grown to become the first global infodemic, the English Task 1 focused on tweets related to COVID-19.

The lab organizers provided a dataset of tweets originating from the early days of the global COVID-19 pandemic and covering a variety of COVID-19-related topics, e.g.,~concerning the number of confirmed cases in different parts of the world, the measures taken by local governments to combat the pandemic, claims about the nature of the virus, etc. The participants were challenged to develop systems to rank a set of input tweets according to their check-worthiness.

Below, we describe the system we developed for the English Task 1, which is an ensemble combining deep contextualized text representations with social context. Our official submission was ranked second-best, and it was almost tied with the winner. We further describe a number of additional experiments and comparisons, which we believe should be useful for future research as they provide some indication about what techniques are effective for the task.

\section{Related Work}
\label{sec:related-work}

The earliest work on claim check-worthiness estimation is the ClaimBuster system \cite{claimbuster}, which was trained on manually annotated US Presidential debates. It used TF.IDF features after discarding words appearing in less than three documents. In addition, for each sentence they calculated a sentiment score, word counts, number of occurrences for 43 Part-of-Speech (POS) tags from the Penn Treebank, as well as the frequency of use of 26 entity types, such as \emph{Person} and \emph{Organization}. They performed feature selection using a random forest and GINI index, and then conducted various experiments using feature subsets passed on to a Support Vector Machine, a Random Forest, and a Na\"{i}ve Bayes classifiers.

In a related line of work, Gencheva \& al. \cite{gencheva-etal-2017-context} created a dataset also based on US Presidential debates, but with annotations from nine fact-checking organizations. Their goal was to mimic the selection strategies of these organizations, and they focused on modeling the context. They reused most of ClaimBuster's features and added the number of named entities within a sentence, the number of words belonging to one of nine possible lexicons, i.e.,~words indicating bias or negative words. For the context, they added features for sentence positioning within a speaker's segment, for the size of the segment a sentence belongs to, as well as for the size of the previous and of the next segments, along with several metadata features. They further trained an LDA topic model \cite{lda} on Presidential debates and used it to extract a distribution over 300 learned topics, which they used as additional features. Finally, they added averaged word2vec word embeddings \cite{word2vec} for each sentence. They trained an SVM classifier as well as a feed-forward neural network with two hidden layers, and ReLU for activation, and they found that the additional context features yielded sizable performance gains.

In follow-up work, Vasileva \& al. \cite{RANLP2019:checkworthiness:multitask} used a multi-task learning neural network that predicts whether a sentence would be selected for fact-checking by each individual fact-checking organization (from a set of nine such organizations), as well as by any of them. 

Yet another follow-up work resulted in the development of the ClaimRank system, which was trained on more data and also included Arabic content \cite{NAACL2018:claimrank}. 
Other related work, also focused on political debates and speeches. For example, Patwari \& al. \cite{Patwari:17} predicted whether a sentence would be selected by a fact-checking organization using a boosting-like model. 

Last but not least, the task was the topic of CLEF in 2018 \cite{clef2018checkthat:task1,clef2018checkthat:overall} and 2019~\cite{clef-checkthat-T1:2019,CheckThat:ECIR2019,clef-checkthat:2019}, where the focus was once again on political debates and speeches, from a single fact-checking organization.

In a slightly different domain, Konstantinovskiy \& al. \cite{konstantinovskiy2018} developed a dataset for check-worthiness estimation consisting of manually annotated sentences from TV debates in the UK. They used InferSent sentence embeddings \cite{infersent}, as well as the number of occurrences of some POS tags, and the number of different named entities within a sentence. They experimented with a number of classifiers such as Logistic Regression, SVMs, Na\"{i}ve Bayes, and Random Forest.

Note that all above systems were trained on speeches and debates, while the task we deal with here is about tweets. 
While we reuse some of the features of these systems, we focus on the first step for finding appropriate data representation, i.e.,~using pre-processing techniques specific for tweets.

\section{Data Pre-processing}
\label{sec:preprocessing}

We applied different pre-processing techniques to the raw tweet text, which we describe in the following subsections. 

\subsection{Default Pre-processing}
We will begin with the description of our \emph{default} pre-processing.
It includes the following processing rules and heuristics:

\paragraph{\textbf{Splitting hashtags into separate words based on UpperCamelCase.}} The UpperCamelCase convention is a way to write multiple words joined together as a single word with the first letter of each of the multiple words capitalized within the joined word. It is a technique commonly used in hashtags in Twitter, e.g.,~the string \texttt{\#TheMoreYouKnow} is composed of four words: \textit{the}, \textit{more}, \textit{you}, and \textit{know}. Such joined-words hashtags are common in Twitter, and thus we attempted to split them, possibly extracting useful text and facilitating the understanding of the tweets. As hashtags can deviate from the standard UpperCamelCase convention, we further added some additional rules to cope with some variations.

\paragraph{\textbf{Unification of the hashtags about COVID-19.}} The dataset contains tweets with hashtags such as \texttt{\#covid19}, \texttt{\#covid\_19}, \texttt{\#Covid2019}. We replaced all such hashtags with the canonical tag \texttt{\#covid-19}. Similarly, we unified the different ways of spelling (or even misspelling) the colloquial term \textit{corona virus}, including hashtags such as \texttt{\#coronavirus}, \texttt{\#corona}, \texttt{\#korona}; we replaced all such wordforms and hashtags with the canonical form \textit{corona virus}.

\paragraph{\textbf{Replacing `\texttt{@}' with `\texttt{user}'.}} In general, the identity of a tweet's author is irrelevant regarding a tweet's check-worthiness, and thus we replaced all user mentions with the special token \textit{user}. However, we preserved the identities of influential public figures and organizations since their status might influence the check-worthiness of the respective tweets. In such cases, we replaced certain usernames with the person's actual name or title. For example, we replaced \texttt{@realDonaldTrump} with \textit{Donald Trump}, \texttt{@VP} with \textit{Vice President}, and \texttt{@WHO} with \textit{World Health Organization}.

\paragraph{\textbf{Replacing URLs with \textit{`url'}}.} The presence of a URL can have an impact on the final check-worthiness label. A classifier might have difficulties differentiating between different target URLs, and thus  we replaced all URLs by the special \textit{url} token.

\paragraph{\textbf{Removing hashtags at the end of tweets.}} Many tweets contained hashtags coming after the meaningful textual statement, and possibly before an included URL, e.g.,~\textit{`This is a scandal! \#Covid-19 \#Upset \#Scandal'}. We observed that such hashtags typically did not bring much information with respect to check-worthiness, and thus we removed them.

\paragraph{\textbf{Expanding shortened quantities.}} We replaced tokens such as \textit{7m} and \textit{12k} with expanded versions such as \textit{7 million} and \textit{12 thousand}, respectively.

\paragraph{\textbf{Removing punctuation marks, except for quotation marks.}} Punctuation does not help much semantically, and thus we removed it. However, we kept quotation marks, which can often indicate the beginning or the ending of a quote, which may be the key point of a claim worth fact-checking.

\subsection{Corona Pre-processing}
    We further applied a special pre-processing hack, which we call \textit{Corona} pre-processing. It replaces the words \textit{covid-19} and \textit{corona virus} with \textit{ebola}, which helps to obtain a more meaningful semantic representation of the input text, as \textit{ebola} is in the vocabulary of pre-trained embeddings and Transformers, while \textit{covid-19} and \textit{corona virus}, which are more recent terms, are not.\\

\subsection{SW+C Pre-processing\label{sec:swc}}
    A third method of pre-processing, which we call \textit{SW+C}, applies the rules of the \textit{Corona} pre-processing, and further removes most stop-words (as listed in the standard NLTK stop-word list). However, it keeps personal and demonstrative pronouns, such as~\textit{he} and \textit{this}, as they provide references that might be important to determine whether a claim is check-worthy.\\

\section{Experiments}
\label{sec:experiments}

We experimented with the above pre-processing techniques and different representations and learning models.
All results reported in Tables~\ref{tab:1}-\ref{tab:3} show the average performance using 5-fold cross-validation on the validation set. 

Our baseline system is an SVM with TF.IDF tokens as features, and it achieved a MAP score of 0.6235. 

\subsection{Experiments with SVM}\label{sec:svmexps}

We used an SVM classifier with word embeddings from GloVe, FastText, Sent2vec, as well as from Transformers. For GloVe and for FastText, we used three different pooling options: mean pooling, max pooling, and TF-IDF pooling. For embeddings from Transformers, we used mean pooling, max pooling, direct use of the CLS token, and WK pooling \cite{wk_pooling}.

For each embedding type and pre-processing technique, we performed randomized search with 1,000 iterations. The space of search hyper-parameters consisted of three different kernel types: linear, polynomial, and RBF, each with an equal probability of being selected. In addition, the regularization parameter $C$ and the kernel coefficient for the RBF and for the polynomial kernels $\gamma$ were sampled from a Gamma distribution with parameters \textalpha =2, \textbeta =1, so that most of our samples are close to the default value of 1 in order to prevent overfitting; however, in some cases, extreme values were also tried. When selecting a polynomial kernel, a degree parameter also needs to be provided. We sampled the degree uniformly with values between 2 and 5. 
The results of the experiments are shown in Table~\ref{tab:1}. 
In addition to reporting the highest achieved MAP score for a given group, we further show the corresponding macro-F1 score.

\begin{table}[tbh]
\caption{\textbf{SVM experiments:} Shown is the highest dev MAP and the corresponding macro-F1 score. The best results for each pre-processing and evaluation measure are underlined, while the best overall result is marked in bold.\label{tab:1}}
\begin{center}
\begin{tabular}{ l@{ }@{ }@{ }c@{ }@{ }@{ }c@{ }@{ }@{ }c@{ }@{ }@{ }c }
\toprule
 & \multicolumn{2}{c}{\bf Corona} & \multicolumn{2}{c}{\bf SW+C} \\
\bf Embedding+Pooling & \bf MAP & \bf macro-F1 & \bf MAP & \bf macro-F1 \\
\midrule
    GloVe mean pooling & 0.7140 & 0.6989 & 0.7239 & 0.7168 \\ 
GloVe max pooling & 0.6709 & 0.6876 & 0.6632 & 0.6510 \\ 
GloVe TF-IDF pooling & 0.6938 & 0.7074 & 0.6974 & 0.7057 \\
\midrule
    FastText mean pooling & 0.7151 & 0.7104 & 0.7286 & 0.7300 \\ 
FastText max pooling & 0.6852 & 0.6968 & 0.6801 & 0.6972 \\ 
FastText TF-IDF pooling & 0.6909 & 0.6932 & 0.6923 & 0.6985 \\
\midrule
Wiki unigrams & 0.7020 & 0.6948 & 0.6563 & 0.6239 \\
Wiki bigrams & 0.6981 & 0.3929 & 0.6883 & 0.6937 \\
Twitter unigrams & 0.7133 & 0.7084 & 0.7092 & 0.7121 \\
Twitter bigrams & 0.7315 & 0.7135 & 0.7141 & 0.7266 \\
Toronto unigrams & 0.6589 & 0.6877 & 0.6473 & 0.6808 \\
Toronto bigrams & 0.6795 & 0.6966 & 0.6767 & 0.6865 \\
\midrule
BERT base & \underline{0.7454} & 0.7300 & \underline{0.7535} & \underline{\bf 0.7656} \\
BERT large & 0.7015 & 0.7228 & 0.7055 & 0.7098 \\
DistilBERT & 0.7357 & \underline{0.7394} & 0.7317 & 0.7022 \\
RoBERTa base & 0.7178 & 0.6877 & 0.7332 & 0.7373 \\
RoBERTa large & 0.7164 & 0.7349 & 0.7254 & 0.7256 \\
\bottomrule
\end{tabular}
\end{center}
\end{table}

We can see in Table~\ref{tab:1} that using mean pooling alongside GloVe embeddings yields marginally better results than using max pooling or TF-IDF pooling. The same is true for the experiments with FastText embeddings, which achieved a slightly higher MAP and macro-F1 scores. The Twitter bigrams Sent2Vec embeddings, combined with a pre-processing of replacing all occurrences of \textit{Covid-19} and \textit{Corona virus} with \textit{Ebola}, managed to outperform the GloVe and the FastText embeddings. Finally, using Transformer embeddings yielded the best overall MAP and macro-F1 scores; note that these results were achieved using WK pooling on the BERT-base embeddings.

\subsection{Experiments with Logistic Regression}\label{sec:expslogregr}

We experimented with Logistic Regression in a setup, similar to that for SVM: we performed randomized search with 1,250 iterations and 5-fold cross-validation. The hyper-parameter search space includes different solvers such as Newton-cg, SAG, LBFGS, SAGA, and Liblinear, and the regularization parameter $C$ was sampled from a Gamma distribution with parameters \textalpha =2, \textbeta =1.

\begin{table}[tbh]
\begin{center}
\caption{\textbf{Logistic regression experiments:} Shown are the highest dev MAP and the corresponding macro-F1 score. The best results for each pre-processing and evaluation measure are underlined, while the best overall result is marked in bold.\label{tab:2}}
\begin{tabular}{ l@{ }@{ }@{ }c@{ }@{ }@{ }c@{ }@{ }@{ }c@{ }@{ }@{ }c }
\toprule
 & \multicolumn{2}{c}{\bf Corona} & \multicolumn{2}{c}{\bf SW+C} \\
\bf Embedding+Pooling & \bf MAP & \bf macro-F1 & \bf MAP & \bf macro-F1 \\
\midrule
GloVe mean pooling & 0.6569 & 0.6755 & 0.6714 & 0.6781 \\
GloVe max pooling & 0.6627 & 0.6783 & 0.6583 & 0.6703 \\
GloVe TF-IDF pooling & 0.6460 & 0.6653 & 0.6502 & 0.6648 \\
\midrule
FastText mean pooling & 0.6536 & 0.6937 & 0.6592 & 0.6759 \\
FastText max pooling & 0.6638 & 0.6594 & 0.6567 & 0.6664 \\
FastText TF-IDF pooling & 0.6431 & 0.6691 & 0.6459 & 0.6715 \\
\midrule
Wiki unigrams & 0.6554 & 0.6871 & 0.6414 & 0.6856 \\
Wiki bigrams & 0.6617 & 0.6601 & 0.6525 & 0.6567 \\
Twitter unigrams & 0.6712 & 0.6748 & 0.6520 & 0.6593 \\
Twitter bigrams & 0.6741 & 0.6739 & 0.6649 & 0.6661 \\
Toronto unigrams & 0.6408 & 0.6540 & 0.6396 & 0.6530 \\
Toronto bigrams & 0.6429 & 0.6448 & 0.6422 & 0.6335 \\
\midrule
BERT base & \underline{0.7494} & \underline{0.7433} & \underline{0.7510} & \underline{\bf 0.7539} \\
BERT large & 0.7255 & 0.7325 & 0.7134 & 0.7195 \\
DistilBERT & 0.7140 & 0.7053 & 0.7156 & 0.7018 \\
RoBERTa base & 0.7488 & 0.7300 & 0.7382 & 0.7381 \\
RoBERTa large & 0.7002 & 0.7201 & 0.7002 & 0.7201 \\
\bottomrule
\end{tabular}
\end{center}
\end{table}

Table~\ref{tab:2} shows the evaluation results.
The highest MAP score for GloVe embeddings is achieved with mean pooling, similarly to SVM. In contrast, the best MAP scores on FastText embeddings are achieved using max pooling (whereas mean pooling was best with SVM). Overall, the results with Logistic Regression and Glove \& FastText embeddings are moderately lower than those with SVM.

The Sent2Vec embedding experiments yielded the best results with Twitter bigram embeddings (as was the case with SVM). However, they only yielded a tiny improvement over the highest scores using GloVe and FastText embeddings.

The results using Transformer embeddings as features for Logistic Regression are similar to the ones with SVM: in both cases, WK pooling with BERT-base worked best, and the MAP scores were also similar. In the case of Logistic Regression, the use of Transformer embeddings considerably improved the MAP scores compared to the use of GloVe, FastText, and Sent2Vec embeddings.

\subsection{Experiments with Transformers}

For Transformers, we used randomized search with 5-fold cross-validation for 20 iterations. We sampled the hyper-parameters uniformly as follows: the number of training epochs from $\{ 2, 3, 4, 5\}$, the batch size from $\{2, 4, 8, 12, 16, 24, 32\}$, and the learning parameter from $\{6.25e-5, 5e-5, 3e-5, 2e-5\}$.

\begin{table}[tbh]
\caption{\textbf{Transformer experiments:} Shown is the highest dev MAP and the corresponding macro-F1 score using Logistic Regression, embeddings from Transformers, and different pooling techniques. The best results for each pre-processing and evaluation measure are underlined, while the best overall result is marked in bold.\label{tab:3}}
\begin{center}
\begin{tabular}{ l@{ }@{ }c@{ }@{ }c@{ }@{ }c@{ }@{ }c@{ }@{ }c@{ }@{ }c }
\toprule
 & \multicolumn{2}{c}{\bf Corona} & \multicolumn{2}{c}{\bf SW+C} & \multicolumn{2}{c}{\bf Default}\\
\bf Embedding+Pooling & \bf MAP & \bf macro-F1 & \bf MAP & \bf macro-F1 & \bf MAP & \bf macro-F1 \\
\midrule
BERT base & 0.7384 & 0.7255 & 0.7311 & \underline{0.7523} & 0.7371 & 0.7558 \\
RoBERTa base & \underline{0.7860} & \underline{0.7672} & \underline{0.7552} & 0.7492 & \underline{0.7854} & \underline{\bf 0.7880} \\
DistilBERT & 0.7433 & 0.7297 & 0.7405 & 0.7411 & 0.7476 & 0.7442 \\
AlBERT base & 0.7340 & 0.7140 & 0.6802 & 0.6471 & 0.6625 & 0.6376 \\
\hline
\end{tabular}
\end{center}
\end{table}

The results are shown in Table~\ref{tab:3}. Among the Transformer models, RoBERTa achieved the highest MAP score by a wide margin and for all pre-processing techniques. The highest score was achieved using the \textit{Corona} pre-processing, which is the best overall result. The hyper-parameters that performed the best used 5 training epochs, a batch size of 8, and a learning rate of 3e-05.

\subsection{Experiments using Tweet Metadata}\label{sec:metadata}

Finally, we used information about the tweet and its author, e.g.,~the number of retweets of the target tweet, the number of friends the tweet's author has, the number of years since the account was created, the presence of a URL in the tweet, etc. 
We used these extra features by concatenating them to the validation set predictions for the best-performing RoBERTa models, which arose from the 5-fold cross-validation. 
The best set of parameters for each RoBERTa model and pre-processing is shown in Table~\ref{tab:robertaparameters}.

As some tweets contain a link to online news articles, we designed features modeling the factuality of reporting of the outlets that published these news articles. For this, we used the manual judgments from Media Bias/Fact Check.\footnote{\url{http://mediabiasfactcheck.com}}

We thus derived the following nine Boolean features:

\begin{itemize}
  \item Is Twitter account verified?
  \item Does the tweet contain a URL?
  \item Does the tweet contain a link to an article published by a news outlet whose factuality of reporting is
  \begin{itemize}
      \item very high?
      \item high?
      \item mostly factual?
      \item mixed?
      \item low?
      \item fake news?
      \item conspiracy?
  \end{itemize}
\end{itemize}

As well as the following three numerical features:

\begin{itemize}
  \item Natural logarithm of the number of times the tweet was retweeted;
  \item Natural logarithm of the number of friends of the Twitter account;
  \item Years since the registration of the Twitter account.
\end{itemize}

We conducted experiments concatenating the RoBERTa predictions to the first nine and also to all twelve metadata features. The results are shown in Tables~\ref{tab:meta9} and~\ref{tab:meta12}, respectively. Once again, we ran randomized search using SVM and Logistic Regression on the new features, while searching in the same hyper-parameter space as described in Sections~\ref{sec:svmexps} and \ref{sec:expslogregr}. We can see that Logistic Regression performed better than SVM. Moreover, while the best MAP scores are identical when using twelve vs. nine metadata features, using all twelve features performed a bit better in terms of macro-F1 score.

\begin{table}[h!]
\caption{\textbf{Metadata experiments:} Pre-processing and hyper-parameters for the best-performing RoBERTa models on the validation set.} \label{tab:robertaparameters}
\begin{center}
\begin{tabular}{ l@{ }@{ }@{ }c@{ }@{ }@{ }c@{ }@{ }@{ }c@{ }@{ }@{ }@{ }@{ }@{ }c }
\toprule
\bf Pre-processing & \bf Training Epochs & \bf Batch Size & \bf Learning Rate & \bf MAP \\
\midrule
Default & 5 & 8 & 3.00e-05 & 0.7900 \\
Corona & 4 & 2 & 2.00e-05 & 0.7875 \\
Default & 4 & 8 & 3.00e-05 & 0.8854 \\
Corona & 4 & 24 & 5.00e-05 & 0.7442 \\
Corona & 5 & 8 & 3.00e-05 & 0.8363 \\
\bottomrule
\end{tabular}
\end{center}
\end{table}

\begin{table}[h!]
\caption{\textbf{Metadata experiments:} Highest MAP and corresponding macro-F1 scores for SVM and Logistic Regression using the RoBERTa prediction and the first nine metadata features.} \label{tab:meta9}
\begin{center}
\begin{tabular}{ l@{ }@{ }@{ }c@{ }@{ }@{ }c }
\toprule
\bf Model & \bf MAP & \bf macro-F1 \\
\midrule
SVM & 0.7994 & 0.7893 \\
Logistic Regression & 0.8017 & 0.7868 \\
\bottomrule
\end{tabular}
\end{center}
\end{table}

\begin{table}[h!]
\begin{center}
\caption{\textbf{Metadata experiments:} Highest MAP and corresponding macro-F1 scores for SVM and Logistic Regression using the RoBERTa prediction and all twelve metadata features.} \label{tab:meta12}
\begin{tabular}{ l@{ }@{ }@{ }c@{ }@{ }@{ }c }
\toprule
 & \bf MAP & \bf macro-F1 \\
\midrule
SVM & 0.7993 & 0.7893 \\
Logistic Regression & 0.8017 & 0.7864 \\
\bottomrule
\end{tabular}
\end{center}
\end{table}

\section{Official Results on the Test Set}
\label{sec:results}

For our primary submission, we used Logistic Regression with all twelve metadata features in addition to RoBERTa predictions, as described in Section~\ref{sec:metadata}.

For our first and second contrastive runs, we used models based solely on RoBERTa. Our contrastive 2 run averages the test set predictions of each of the trained RoBERTa models with identical hyper-parameters, whereas our contrastive 1 run averages the predictions of the best-performing RoBERTa models on the corresponding validation sets. For the contrastive 1 run, we did not pose the restriction of having identical hyper-parameters. 

The official evaluation results are shown in Table~\ref{tab:testresults}.
We can see that our primary run achieved a MAP score of 0.8034, which puts us at second place, falling behind the winner by 0.003 points absolute only, i.e.,~we are practically tied for the first place. Our primary run achieved a higher MAP score on the test set than on the validation sets, which is a sign of a model that is capable of generalizing well and not overfitting. 
Our first and second contrastive runs also achieved high MAP scores on the test set, but were slightly worse.

\begin{table}[tbh]
\begin{center}
\caption{\textbf{Official results on the test set.} Shown are the results for our primary and constrastive submissions. \label{tab:testresults}}
\begin{tabular}{ l@{ }@{ }@{ }c@{ }@{ }@{ }c@{ }@{ }@{ }c@{ }@{ }@{ }c@{ }@{ }@{ }c@{ }@{ }@{ }@{ }@{ }@{ }c@{ }@{ }@{ }c@{ }@{ }@{ }c }
\toprule
\bf Runs & \bf MAP & \bf R-Pr & \bf P@1 & \bf P@3 & \bf P@5 & \bf P@10 & \bf P@20 & \bf P@30 \\
\midrule
Primary & 0.8034 & 0.6500 & 1.0000 & 1.0000 & 1.0000 & 1.0000 & 0.9500 & 0.7400 \\
Contrastive 1 & 0.7988 & 0.6500 & 1.0000 & 1.0000 & 1.0000 & 1.0000 & 0.9500 & 0.7400 \\
Contrastive 2 & 0.7809 & 0.6667 & 1.0000 & 1.0000 & 1.0000 & 1.0000 & 0.8500 & 0.6800 \\
\bottomrule
\end{tabular}
\end{center}
\end{table}
\section{Conclusion and Future Work}
\label{sec:conclusion}

We have described our system, Team\_Alex, for detecting check-worthy tweets about COVID-19, which we developed for the English version of CLEF-2020 CheckThat! \mbox{Task 1}. It is based on an ensemble combining deep contextualized text representations with social context, as well as advanced pre-processing. Our system was ranked second with a MAP score of 0.8034, which is almost tied with the wining system, lagging behind by just 0.003 MAP points absolute. 

We further described a number of additional experiments and comparisons, which we believe should be useful for future research as they provide some indication about what techniques are effective for the task.

In future work, we plan to experiment with more pre-processing techniques, with better modeling the social context, as well as with some newer Transformer models such as T5 \cite{t5} and Electra \cite{electra}.

\section*{Acknowledgments}

This research is part of the Tanbih project,\footnote{\url{http://tanbih.qcri.org}} developed at the Qatar Computing Research Institute, HBKU, which aims to limit the effect of ``fake news'', propaganda, and media bias by making users aware of what they are reading, thus promoting media literacy and critical thinking.

This research is also partially supported by the National Scientific Program ``ICTinSES'', financed by the Bulgarian Ministry of Education and Science and Project UNITe BG05M2OP001-1.001-0004, funded by the OP ``Science and Education for Smart Growth'' and the EU via the ESI Funds.

\bibliographystyle{splncs04}
\bibliography{ref}

\end{document}